\documentclass[11pt,letterpaper]{article}
\usepackage{emnlp2017}
\usepackage{times}
\usepackage{latexsym}

\usepackage{tikz}
\usepackage{tikz-qtree}
\usepackage{float}

\usepackage{mathpartir}
\usepackage{proof}
\usepackage{xspace}
\usepackage{multirow}

\newcommand{\shift}{\ensuremath{\mathsf{sh}}\xspace}
\newcommand{\comb}{\ensuremath{\mathsf{comb}}\xspace}
\newcommand{\labelx}{\ensuremath{\mathsf{label}_X}\xspace}
\newcommand{\nolabel}{\ensuremath{\mathsf{nolabel}}\xspace}

\newcommand{\score}{\ensuremath{\mathit{sc}}\xspace}
\newcommand{\scoring}[2]{\ensuremath{\score_{#1}({#2})}\xspace}
\newcommand{\scoreshift}[1]{\ensuremath{\scoring{\shift}{#1}}\xspace}
\newcommand{\scorecomb}[1]{\ensuremath{\scoring{\comb}{#1}}\xspace}
\newcommand{\scorelabelx}[1]{\ensuremath{\scoring{\labelx}{#1}}\xspace}
\newcommand{\scorenolabel}[1]{\ensuremath{\scoring{\nolabel}{#1}}\xspace}

\newcommand*\trapezoid{\resizebox{0.4cm}{0.26cm}{\includegraphics{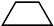}}}
\newcommand*\trivialtrapezoid{\resizebox{0.23cm}{0.26cm}{\includegraphics{trapezoid.pdf}}\,}

\newcommand*\longtrapezoid[1]{\resizebox{0.8cm}{0.26cm}{\trapezoid}\hspace{-0.45cm}{\raisebox{0.1cm}{$_{#1}$}}\hspace{0.35cm}}
\newcommand*\shorttrapezoid[1]{\resizebox{0.5cm}{0.26cm}{\trapezoid}\hspace{-0.3cm}{\raisebox{0.1cm}{$_{#1}$}}\hspace{0.2cm}}

\emnlpfinalcopy

\def\emnlppaperid{***}

\makeatletter
\def\blfootnote{\gdef\@thefnmark{}\@footnotetext}
\makeatother

\newcommand{\tuple}[1]{\ensuremath{\langle {#1} \rangle}}

\title{Joint Syntacto-Discourse Parsing and the Syntacto-Discourse Treebank
\thanks{\ \ \ The source code and the joint treebank are available
  at \url{https://github.com/kaayy/josydipa}.}}

\author{Kai Zhao$^\dagger$ \and Liang Huang \\
  School of Electrical Engineering and Computer Science\\
  Oregon State University\\
  Corvallis, Oregon, USA\\
  {\tt \{kzhao.hf, liang.huang.sh\}@gmail.com}
}

\date{}

\begin{document}

\maketitle

\begin{abstract}
  Discourse parsing has long been treated as a stand-alone problem
  independent from constituency or dependency parsing. 
  Most attempts at this problem are pipelined rather than end-to-end, 
  sophisticated, and not self-contained:
  they assume gold-standard text segmentations 
  (Elementary Discourse Units), 
  and use external parsers for syntactic features.
  In this paper we propose the first end-to-end discourse parser
  that jointly parses in both syntax and discourse levels,
  as well as the first syntacto-discourse treebank by integrating 
  the Penn Treebank with the RST Treebank.
  Built upon our recent span-based constituency parser,
  this joint syntacto-discourse parser requires no preprocessing whatsoever (such as segmentation or feature
  extraction), 
achieves the state-of-the-art end-to-end discourse parsing accuracy.
\end{abstract}

\blfootnote{$^\dagger$ Current address: Google Inc., 
New York, NY, USA.}

\section{Introduction}\label{sec:intro}

Distinguishing the semantic relations between segments in a document
 can be greatly beneficial to many high-level NLP tasks,
such as summarization \cite{louis2010discourse,yoshida2014dependency}, sentiment analysis \cite{voll2007not,somasundaran2009supervised,bhatia2015better},
question answering \cite{ferrucci2010building,jansen2014discourse}, and textual quality evaluation \cite{tetreault2013holistic,li2016neural}.

There has been a variety of research on discourse parsing 
\cite{marcu2000rhetorical,soricut2003sentence,pardo2008development,hernault2010hilda,da2012symbolic,joty2013combining,joty2014discriminative,feng2014linear,ji2014representation,li2014recursive,li2014text,heilman2015fast,wang2017two}.
But most of them suffer from the following limitations:
\begin{enumerate}
 \item {\em pipelined rather than end-to-end}: they assume pre-segmented discourse, and worse yet, use gold-standard segmentations, except \newcite{hernault2010hilda};        
 \item {\em not self-contained}: they rely on external syntactic parsers and pretrained word vectors; 
 \item {\em complicated}: they design sophisticated features, including those from parse-trees.  
\end{enumerate}

We argue for the first time that discourse parsing should be viewed 
as an extension of, and be performed in conjunction with, constituency parsing.
We propose the first {\it joint syntacto-discourse treebank}, 
by 
unifying constituency and discourse tree representations.
Based on this, we propose the first
{\it end-to-end} incremental parser
that jointly parses at both constituency and discourse levels.
Our algorithm
builds up on the span-based parser \cite{cross2016incremental};
it employs the strong generalization power of bi-directional LSTMs,
and parses efficiently and robustly with an extremely simple span-based feature set
that does not use any tree structure information.

We make the following contributions:

\begin{figure*}[h]
\centering
\tikzset{sibling distance=0.1cm, level distance=0.9cm}
\resizebox{\textwidth}{!}{
\begin{tikzpicture}
\Tree[.{} [.\node(a){$\circ$}; 
                               {Costa Rica had been negotiating with U.S. banks} ] 
          [.\node(b){$\bullet$}; [.\node(c){$\bullet$}; 
                                                        {but the debt plan was rushed to completion} ] 
                                 [.\node(d){$\circ$}; 
                                                      {in order to be announced at the meeting} ] ] ]
\draw[->, dashed] (a)-- node[above] {Background} ++(b);
\draw[->, dashed] (d)-- node[above] {Purpose} ++(c);
\end{tikzpicture}
}\\
(a) A discourse tree with 3 EDUs ($\bullet$: nucleas; $\circ$: satellite) in the RST treebank \cite{marcu2000theory}\\[0.2cm]
\resizebox{\textwidth}{!}{
\begin{tikzpicture}
\tikzset{sibling distance=-0.15cm, level distance=.7cm}
\Tree[.{Background$\rightarrow$} [.S [.NP [.NNP Costa ] [.NNP Rica ] ]
                                     [.VP [.VBD had ]
                                          [.VP [.VBN been ] [.VP [.VBG negotiating ] 
                                                                 [.PP [.IN with ] 
                                                                      [.NP [.NNP U.S. ] [.NNS banks ] ] ]
                                                            ] ] ] ]
          [.{$\leftarrow$Purpose} [.S [.CC but ] 
                                      [.S [.NP [.DT the ] [.NN debt ] [.NN plan ]  ]
                                      [.VP [.VBD was ] 
                                           [.VP [.VBN rushed ] 
                                                [.PP [.TO to ] 
                                                     [.NP [.NN completion ]  ] ] ] ] ] ]
                                  [.SBAR [.IN in ] [.NN order ] 
                                                  [.S [.VP [.TO to ] 
                                                           [.VP [.VB be ] 
                                                                [.VP [.VBN announced ] 
                                                                     [.PP [.IN at ] 
                                                                          [.NP [.DT the ] [.NN meeting ]
                                                                          ] ] ] ] ] ] ] ] ]
\end{tikzpicture}
}\\[-0.3cm]
(b) The corresponding RST-PTB tree (our work)
\caption{Examples of the RST discourse treebank and our syntacto-discourse treebank (PTB-RST).\label{fig:firstexample}}
\end{figure*}
 
\begin{enumerate}
	\item We develop a combined representation of constituency and discourse trees to facilitate 
	parsing at both levels without explicit conversion mechanism.
        Using this representation, we build and release a joint treebank based on the Penn
        Treebank \cite{marcus1993building} and RST Treebank \cite{marcu2000rhetorical,marcu2000theory}
        (Section~\ref{sec:combined}).

	\item We propose a novel joint parser that parses at both constituency and discourse levels.
	Our parser performs discourse parsing in an end-to-end manner, which greatly reduces
	the efforts required in preprocessing the text for segmentation and feature extraction,
	and, to our best knowledge, is the first end-to-end discourse parser in literature (Section~\ref{sec:parsing}).

	\item Even though it simultaneously performs constituency parsing,
        our parser does {\em not} use 
        any explicit syntactic feature, 
        nor does it need any binarization of discourse trees,
        thanks to the powerful span-based framework of \newcite{cross2016incremental}
        (Section~\ref{sec:parsing}).

	\item Empirically, our end-to-end parser outperforms the existing pipelined 
	discourse parsing efforts. When the gold EDUs are provided, our parser is also competitive to other
	existing approaches with sophisticated features (Section~\ref{sec:exps}).
\end{enumerate}
 
\section{Combined Representation \& Treebank}\label{sec:combined}

We first briefly review the discourse structures in Rhetorical Structure Theory 
\cite{mann1988rhetorical},
and then discuss how to unify discourse and constituency trees, 
which gives rise to our syntacto-discourse treebank PTB-RST.

\begin{figure*}[ht]
\centering
\resizebox{\textwidth}{!}{
\begin{tikzpicture}
\tikzset{sibling distance=0.05cm, level distance=.8cm}
\Tree[.{} [.\node(a){$\bullet$}; 
                                 {The metals sector outgained other industry groups.} ] 
          [.\node(b){$\circ$}; [.\node(c){$\bullet$}; 
                                                      {Hecla Mining rose $\frac{5}{8}$ to 14;} ] 
                               [.\node(d){$\bullet$}; 
                                                      {Battle Mountain Gold climbed $\frac{3}{4}$ to 16$\frac{3}{4}$;} ] 
                               [.\node(e){$\bullet$}; 
                                                      {and ASA Ltd.~jumped 3$\frac{5}{8}$ to 49$\frac{5}{8}$.} ] 
          ] 
     ]
\draw[->, dashed] (b)-- node[above] {Elaboration} ++(a);
\draw[-, dashed] (c)-- node[above] {List} ++(e);

\end{tikzpicture}
}\\[0.1cm]

\resizebox{\textwidth}{!}{
\begin{tikzpicture}
\tikzset{sibling distance=0.05cm, level distance=.7cm}
\Tree [.{$\leftarrow$Elaboration} 
        [.S 
          [.NP \edge[roof]; {The metals sector} ]
          [.VP 
            [.VBD outgained ]
            [.NP \edge[roof]; {other industry groups.} ]
          ]
        ]
        [.{List}
          [.S 
            [.NP \edge[roof]; {Hecla Mining} ]
            [.VP \edge[roof]; {rose $\frac{5}{8}$ to 14;} ]
          ]
          [.S 
            [.NP \edge[roof]; {Battle Mountain Gold} ]
            [.VP \edge[roof]; {climbed $\frac{3}{4}$ to 16$\frac{3}{4}$;} ]
          ]
          [.S 
            [.CC and ]
            [.S \edge[roof]; {ASA Ltd.~jumped 3$\frac{5}{8}$ to 49$\frac{5}{8}$.} ]
          ]
        ]
      ]
\end{tikzpicture}
}\\[-0.3cm] 
\caption{
Another example of RST vs.~PTB-RST, demonstrating
a discourse tree over two sentences and a non-binary relation (List).
The lower levels of the PTB-RST tree are collapsed due to space contraints.
\label{fig:secondexample}}
\vspace{-0.7cm}
\end{figure*}
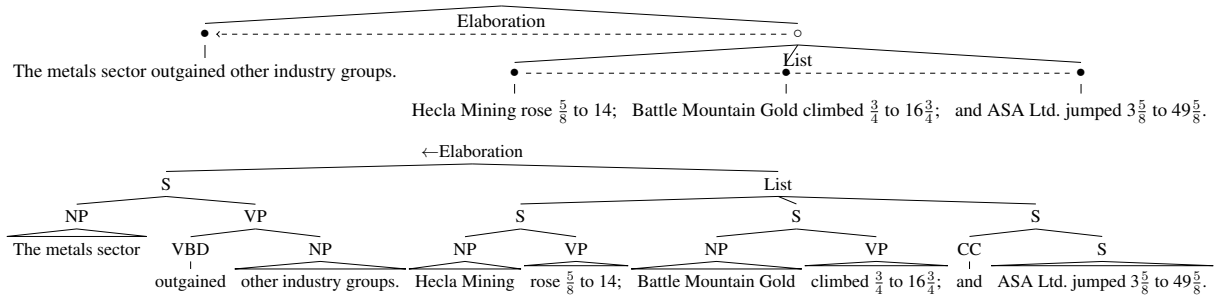

\subsection{Review: RST Discourse Structures}

In an RST 
discourse tree,
there are two types of branchings. 
Most of the internal 
tree nodes are binary branching, with one {\it nucleus} child 
containing the core semantic meaning of the current node, 
and one {\it satellite} child semantically decorating 
the nucleus. Like dependency labels,
there is a {\em relation} annotated 
between each satellite-nucleus pair, 
such as ``Background'' or ``Purpose''.
Figure~\ref{fig:firstexample}(a) shows an example RST tree.
There are also non-binary-branching internal nodes whose children are conjunctions, e.g., a ``List'' of semantically similar
EDUs (which are all nucleus nodes); see Figure~\ref{fig:secondexample}(a) for an example.

\subsection{Syntacto-Discourse Representation}

It is widely recognized that lower-level lexical and syntactic information can greatly help 
determining both the boundaries of the EDUs 
(i.e., discourse segmentation) \cite{bach2012reranking} 
as well as the semantic relations between EDUs 
\cite{soricut2003sentence,hernault2010hilda,joty2014discriminative,feng2014linear,ji2014representation,li2014recursive,heilman2015fast}.
While these previous approaches 
rely
on pre-trained tools to provide 
both EDU segmentation and
intra-EDU syntactic parse trees,
we instead propose to directly determine 
the low-level segmentations,
the syntactic parses,
and the high-level discourse parses
using a single joint parser.
This parser is trained on the combined trees of constituency 
and discourse structures.

We first convert an RST tree to a format similar to those constituency trees in the Penn Treebank \cite{marcus1993building}.
For each binary branching node with a nucleus child and a satellite child, we use the relation as the label of the converted
parent node. The nucleus/satellite relation, along with the direction
(either $\leftarrow$ or $\rightarrow$, pointing from satellite to nucleus) 
is then used as the label. 
For example, at the top level in Figure~\ref{fig:secondexample},
we convert 
\[
\resizebox{4cm}{!}{
\begin{tikzpicture}[scale=.8]
\tikzset{sibling distance=4.5cm, level distance=.8cm}
\Tree[.{} [.\node(a){$\bullet$}; 
                                 {$\ldots$} ]
          [.\node(b){$\circ$};  {$\ldots$} ] 
          ] 
     ]
\draw[->, dashed] (b)-- node[above] {Elaboration} ++(a);
\end{tikzpicture}
}
\mbox{\ to \ }
\begin{tikzpicture}[level distance=20,scale=0.8, sibling distance=2cm]
\Tree [.{$\leftarrow$Elaboration} {$\ldots$} {$\ldots$} ]
\end{tikzpicture}
\]
For a conjunctive branch (e.g.~``List''), we simply use the relation as the label of the converted node. 

After converting an RST tree into the constituency tree format, 
we then replace each leaf node (i.e., EDU) with the corresponding syntactic (sub)tree from PTB. 
Given that the sentences in the RST Treebank \cite{marcu2000theory} 
is a subset of that of PTB,
we can always find the corresponding constituency subtrees for each EDU leaf node.
In most cases, each EDU corresponds to one single (sub)tree in PTB, 
since the discourse boundaries generally do not conflict with constituencies. 
In other cases, one EDU node may correspond to multiple subtrees in PTB,
and for these EDUs we 
use the lowest common ancestor of those subtrees in the PTB as the label of that EDU in the converted tree.
E.g., if C--D is one EDU in the PTB tree
\tikzset{sibling distance=0.05cm, level distance=.6cm}
{\small \Tree[.A B C D ]} it might be converted to {\small \Tree[.{Purpose$\rightarrow$} B [.A C D ] ]}
if the relation annonated in RST is B $\stackrel{\mbox{\small Purpose}}{\longrightarrow}$ C--D.

Figures~\ref{fig:firstexample}--\ref{fig:secondexample} are two examples of discourse trees and their combined syntacto-discourse trees.

\begin{figure}
\centering
\includegraphics[width=0.33\textwidth,height=2.6cm]{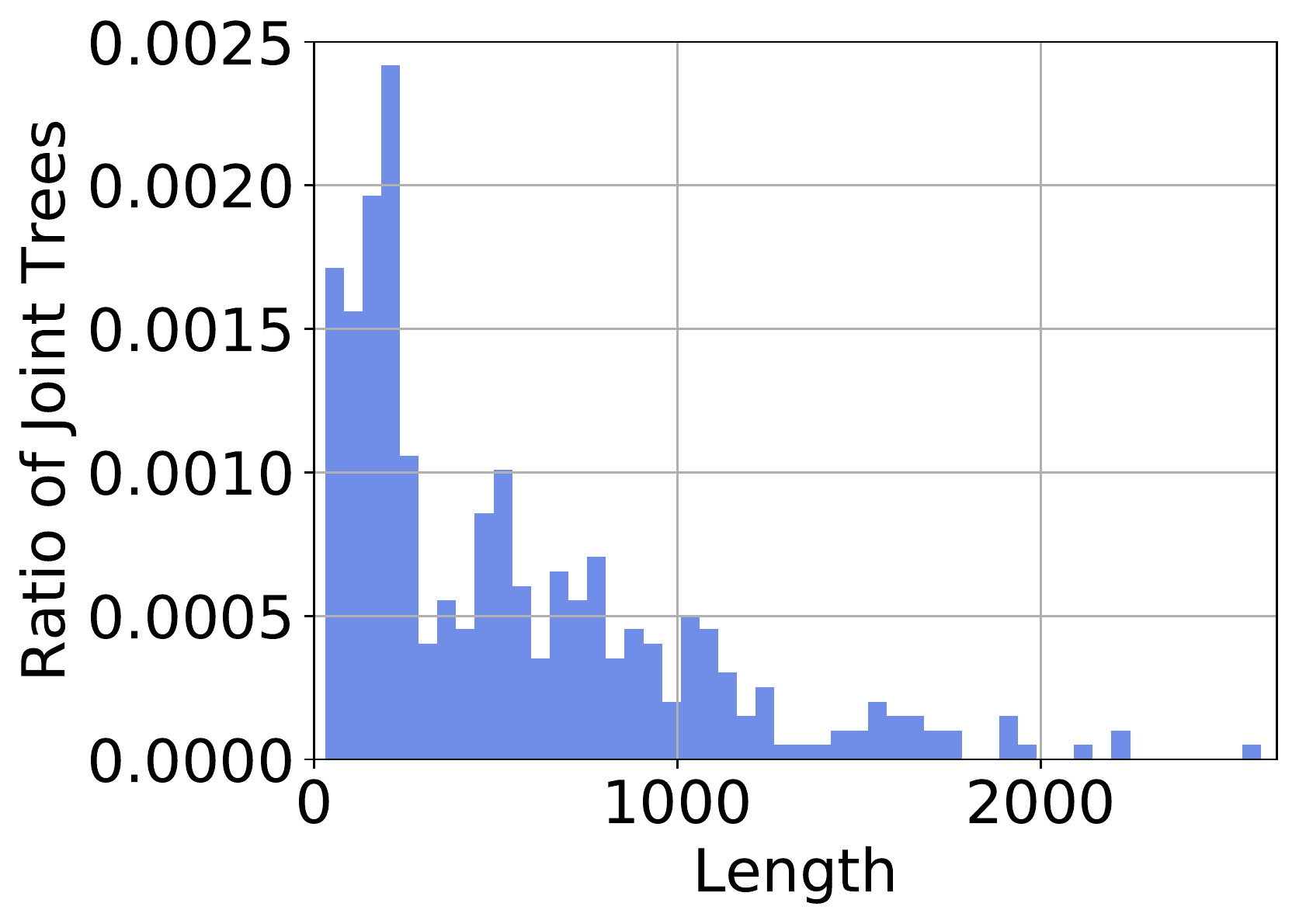}\\[-0.3cm]
\caption{PTB-RST: length distribution (\# tokens).\label{fig:treelen}}
\end{figure}

\subsection{Joint PTB-RST Treebank}

Using the conversion strategy described above 
we build the first joint syntacto-discourse
treebank based on the Penn Treebank 
and RST Treebank. 
This PTB-RST treebank is released as a set of tools to generate the joint trees 
given Penn Treebank and RST Treebank data.
During the alignment between the RST trees and the PTB trees,
we only keep the common parts of the two trees.

We follow the standard training/testing split of the RST Treebank.
In the training set, there are 347 joint trees with 
a total of 17,837 tokens,
and the lengths of the discourses range from 30 to 2,199 tokens.
In the test set, there are 38 joint trees
with a total of 4,819 tokens,
and the lengths vary from 45 to 2,607.
Figure~\ref{fig:treelen} shows the distribution of the discourse lengths
over the whole dataset,
which on average is about 2x of PTB sentence length,
but longest ones are about 10x the longest lengths in the Treebank.

\section{Joint Syntacto-Discourse Parsing}\label{sec:parsing}

Given the combined syntacto-discourse treebank, we now propose a joint parser
that can perform end-to-end discourse segmentation and parsing.

\subsection{Extending Span-based Parsing}

\begin{figure}
\hspace{-0.4cm}
\resizebox{0.53\textwidth}{!}{
  \begin{tabular}{cl}
   input      & $w_0 \ldots w_{n-1}$ \\[0.1in]
    axiom    & \tuple{\ _{-1} \trivialtrapezoid _0}: $(0, \emptyset)$ \quad    goal  \quad \tuple{\ _{-1} \trivialtrapezoid _0 \trapezoid _n}: $(\_, t)$ \\[0.1in]
    \shift   & $\inferrule{\tuple{ ...\ _i \trapezoid _j}: (c, t)}
                          {\tuple{... \ _i \trapezoid _j \shorttrapezoid{j} _{j+1}}: (c+\scoreshift{i, j}, t)}$ $j<n$ \\[0.25in]
    \comb    & $\inferrule{\tuple{ ... \ _i \trapezoid _k \trapezoid _j}: (c, t) }
                          {\tuple{... \ _i \longtrapezoid{k} _j}: (c + \scorecomb{i, k, j}, t)}$ \\[0.25in]
    \labelx  & $\inferrule{\tuple{... \ _i \longtrapezoid{k} _j}: (c, t)}
                          {\tuple{ ... \ _i \trapezoid _j}: (c+\scorelabelx{i, k, j}, t\cup\{_i X_j\})} $ \\[0.25in]
    \nolabel & $\inferrule{\tuple{... \ _i \longtrapezoid{k} _j}: (c, t)}
                          {\tuple{ ... \ _i \trapezoid _j}: (c +\scorenolabel{i,k,j}, t)}$ 
  \end{tabular}
  }
\caption{Deductive system for joint syntactic and discourse parsing. 
$\scoreshift{\cdot, \cdot}$, $\scorecomb{\cdot, \cdot, \cdot}$, 
$\scorelabelx{\cdot, \cdot, \cdot}$, and $\scorenolabel{\cdot, \cdot, \cdot}$ are 
scoring functions evaluated in the neural network.}
\label{fig:sr}
\end{figure}

As mentioned above, the input sequences are substantially longer than PTB parsing, 
so we choose linear-time parsing,
by adapting a popular greedy constituency parser,
the span-based constituency parser of \newcite{cross2016incremental}.

As in span-based parsing, at each step, we maintain a  
a stack of spans.
Notice that in conventional incremental parsing, the stack stores the subtrees constructed so far,
but in span-based constituency parsing, the stack only stores the boundaries of subtrees, 
which are just a list of indices $... _i \trapezoid _k \trapezoid _j$. 
In other words, quite shockingly, no tree structure is represented anywhere in the parser.
Please refer \newcite{cross2016incremental} for details. 

Similar to span-based constituency parsing,
we alternate between structural (either shift or combine) and label (\labelx or \nolabel) actions in an odd-even fashion.
But different from \newcite{cross2016incremental}, after a structural action, we choose to keep the last 
branching point $k$, i.e.,  $ _i \longtrapezoid{k} _j$ (mostly for combine, but also trivially for shift). This is because in our parsing mechanism,
the discourse relation between two EDUs is actually determined after the previous combine action.
We need to keep the splitting point to clearly find the spans of the two EDUs to determine their relations.
This midpoint $k$ disappears after a label action;
therefore we can use the shape of the last span on the stack (whether it contains the split point,
i.e., $ _i \longtrapezoid{k} _j$ or $_i \trapezoid _j$)
to determine the parity of the step and thus no longer need to carry the step $z$ in the state as in \newcite{cross2016incremental}.

The \nolabel action makes the binarization of the discourse/constituency tree unnecessary,
because \nolabel actually combines the top two spans on the stack $\sigma$ into one span,
but without annotating the new span a label.
This greatly simplifies the pre-processing and post-processing efforts needed.

\begin{table}
  \begin{center}
  \resizebox{0.35\textwidth}{!}{
  \begin{tabular}{c|c|c|c}
    & Prec. & Recall & F1 \\
    \hline
    \hline
    Constituency & 87.6 & 86.9 & 87.2 \\
    Discourse & 46.5 & 40.2 & 43.0 \\
    \hline
    Overall & 83.5 & 81.6 & 82.5
  \end{tabular}
  }
  \end{center}
  \caption{Accuracies on PTB-RST at constituency and discourse levels. 
    \label{tab:internal-acc}
  }
  \vspace{-0.4cm}
\end{table}

\begin{table*}
\begin{center}
\resizebox{0.99\textwidth}{!}{
\begin{tabular}{c|c|c|c|ccc}
& description & syntactic feats. & segmentation & structure & +nuclearity & +relation \\
\hline\hline
\newcite{bach2012reranking} &segmentation only & Stanford & 95.1 & - & - & -\\
\hline
\newcite{hernault2010hilda} & end-to-end pipeline & Penn Treebank & 94.0 & 72.3 & 59.1 & 47.3 \\
\hline
\multicolumn{2}{c|}{joint syntactic \& discourse parsing } & - & {\bf 95.4} & {\bf 78.8} & {\bf 65.0} & {\bf 52.2}\\
\end{tabular}
}
\end{center}
\caption{F1 scores of end-to-end systems. ``+nuclearity'' 
indicates scoring of tree structures with nuclearity included. 
``+relation'' has both nuclearity and relation included (e.g., $\leftarrow$Elaboration).
\label{tab:results} }
\vspace{-0.1cm}
\end{table*}

 \begin{table*}
\begin{center}
\resizebox{0.99\textwidth}{!}{
\begin{tabular}{c|c|c|ccc}
& & syntactic feats &  structure & +nuclearity & +relation \\
\hline
\multicolumn{2}{c|}{human annotation \cite{ji2014representation}} & - & 88.7 & 77.7 & 65.8\\
\hline\hline
\multirow{5}{*}{sparse} &\newcite{hernault2010hilda} & Penn Treebank&  83.0 & 68.4 & 54.8\\
&\newcite{joty2013combining} & Charniak (retrained) &82.7 & 68.4 & 55.7\\
&\newcite{joty2014discriminative} &  Charniak (retrained) & - & - & 57.3\\
&\newcite{feng2014linear} &  Stanford &85.7 & 71.0 & 58.2\\
&\newcite{heilman2015fast} &  ZPar (retraied)&83.5 & 68.1 & 55.1\\
&\newcite{wang2017two} & Stanford & {\bf 86.0} & {\bf 72.4} & 59.7 \\
\hline
\multirow{4}{*}{neural} &\newcite{li2014recursive} & \multirow{2}{*}{Stanford} &82.4 & 69.2 & 56.8 \\
&+ sparse features &  &84.0 & 70.8 & 58.6 \\
&\newcite{ji2014representation} & \multirow{2}{*}{MALT} & 80.5 & 68.6 & 58.3\\
&+ sparse features &  &81.6 & 71.1 & {\bf 61.8}\\
\hline\hline
 &span-based discourse parsing & - & 84.2 & 67.7 & 56.0
\end{tabular}
}
\end{center}
\caption{Experiments using gold segmentations.
The column of ``syntactic feats'' shows how the syntactic features
are calculated in the corresponding systems. Note that our parser
predicts solely based on the span features from bi-directionaly LSTM,
instead of any explicitly designed syntactic features.
\label{tab:results-gold} }
\end{table*}
 
\subsection{Recurrent Neural Models and Training}

The scoring functions in the deductive system (Figure~\ref{fig:sr})
are calculated by an underlying neural model,
which is similar to the bi-directional LSTM model
in \newcite{cross2016incremental} that evaluates based on span boundary features.
Again, it is important to note that no discourse or syntactic tree structures are represented in the features.

During the decoding time, 
a document is firstl passed into a two-layer bi-directional LSTM
model, then the outputs at each text position of the two layers of the bi-directional
LSTMs are concatenated as the positional features.
The spans at each parsing step can be represented as the feature vectors at the boundaries.
The span features are then passed into fully connected networks with softmax to calculate the 
likelihood of performing the corresponding action or marking the corresponding label.

We use the ``training with exploration'' strategy \cite{goldberg2013training}
and the dynamic oracle mechanism described in \newcite{cross2016incremental}
to make sure the model can handle unseen parsing configurations properly.

\section{Empirical Results}\label{sec:exps}

We use the treebank described in Section~\ref{sec:combined}
for empirical evaluation.
We randomly choose 30 documents from the training set as the development set.

We tune the hyperparameters of the neural model on the development set.
For most of the hyperparameters we settle with the same values
suggested by \newcite{cross2016incremental}.
To alleviate the overfitting problem for training on the relative small
RST Treebank, we use a dropout of 0.5.

One particular hyperparameter is that we use a value $\beta$ to
balance the chances between training following the exploration (i.e., the best action
chosen by the neural model)
and following the correct path provided by the dynamic oracle.
We find that $\beta=0.8$, i.e., following the dynamic oracle
with a probability of 0.8, achieves the best performance.
One explanation for this high chance to follow the oracle is that, since our combined trees are significantly larger 
than the constituency trees in Penn Treebank,
lower $\beta$ makes the parsing easily divert into wrong trails that are difficult to learn from.

Since our parser essentially performs both constituency parsing task and discourse parsing task.
We also evaluate the performances on sentence constituency level and
discourse level separately. The result is shown in Table~\ref{tab:internal-acc}.
Note that in constituency level, the accuracy is not directly comparable with the accuracy
reported in \newcite{cross2016incremental}, since:
a) our parser is trained on a much smaller dataset (RST Treebank
is about 1/6 of Penn Treebank); b) the parser is trained to optimize
the discourse-level accuracy.

Table~\ref{tab:results} shows that, 
in the perspective of end-to-end discourse parsing,
our parser first outperforms the state-of-the-art segmentator
of \newcite{bach2012reranking},
and furthermore, in end-to-end parsing, the superiority
of our parser is more pronounced comparing to
the previously best parser of \newcite{hernault2010hilda}.

On the other hand, the majority of the conventional discourse parsers are not end-to-end:
they rely on gold EDU segmentations and pre-trained tools like Stanford parsers to generate features.
We perform an experiment to compare the performance of our parser
with them given the gold EDU segments (Table~\ref{tab:results-gold}).
Note that most of these parsers do not handle multi-branching discourse nodes and are trained and evaluated
on binarized discourse trees \cite{feng2014linear,li2014recursive,li2014text,ji2014representation,heilman2015fast}, so their performances are actually not directly comparable to
the results we reported.

\section{Conclusion}
We have presented a neural-based incremental parser that can jointly parse
at both constituency and discourse levels.
To our best knowledge, this is the first end-to-end parser for discourse parsing task.
Our parser achieves the state-of-the-art performance in end-to-end parsing,
and unlike previous approaches, needs little pre-processing effort.

\section*{Acknowledgments}
{
We thank the anonymous reviewers for helpful comments.
We are also grateful to Mingbo Ma, James Cross, and Dezhong Deng for suggestions.
This work is supported in part by
NSF IIS-1656051,
DARPA N66001-17-2-4030 (XAI),
a Google Faculty Research Award,
and HP.
}
\bibliography{emnlp2017}
\bibliographystyle{emnlp_natbib}

\end{document}